\documentclass[10pt,twocolumn,letterpaper]{article}
\usepackage[pagenumbers]{cvpr}
\usepackage{cvpr}              

\usepackage{graphicx}
\usepackage{amsmath}
\usepackage{amssymb}
\usepackage{booktabs}
\usepackage{setspace}
\usepackage{cuted}
\usepackage[pagebackref,breaklinks,colorlinks]{hyperref}

\usepackage[capitalize]{cleveref}
\crefname{section}{Sec.}{Secs.}
\Crefname{section}{Section}{Sections}
\Crefname{table}{Table}{Tables}
\crefname{table}{Tab.}{Tabs.}

\def\cvprPaperID{*****} 
\def\confName{CVPR}
\def\confYear{2022}

\begin{document}
\title{SEREP: Semantic Facial Expression Representation \\for Robust In-the-Wild Capture and Retargeting}
\author
{\parbox{\textwidth}{\centering Arthur Josi $^{1, 2}$\thanks{Equal contribution}\;\;
                                Luiz Gustavo Hafemann$^{2*}$\;\;
                                Abdallah Dib$^{2}$\;\;
                                Emeline Got$^2$\;\;
                                Rafael M. O. Cruz$^1$\;\;
                                Marc-André Carbonneau$^2$
        }
        \\
        \\
{\parbox{\textwidth}{\centering  Ecole de Technologie Supérieure$^1$\;\;\; Ubisoft LaForge$^2$\;\;\;
       }
}
\vspace{-20px}
}

\maketitle



\begin{strip}\centering
    \centering
    \includegraphics[width=1\linewidth]{figures/teaser_small.pdf}
    \captionof{figure}{SEREP extracts facial expressions from in-the-wild monocular images and applies them to a given neutral mesh, enabling accurate expression reconstruction and identity-preserving retargeting.}
    \label{fig:enter-label}
\end{strip}

\begin{abstract}

Monocular facial performance capture in-the-wild is challenging due to varied capture conditions, face shapes, and expressions. Most current methods rely on linear 3D Morphable Models, which represent facial expressions independently of identity at the vertex displacement level. We propose SEREP (Semantic Expression Representation), a model that disentangles expression from identity at the semantic level. We start by learning an expression representation from high-quality 3D data of unpaired facial expressions. Then, we train a model to predict expression from monocular images relying on a novel semi-supervised scheme using low quality synthetic data. In addition, we introduce MultiREX, a benchmark addressing the lack of evaluation resources for the expression capture task. Our experiments show that SEREP outperforms state-of-the-art methods, capturing challenging expressions and transferring them to new identities. Results in the \href{https://ubisoft-laforge.github.io/character/serep/}{project page}.

\end{abstract}    
\section{Introduction}

Facial performance capture is essential for applications where users interact with 3D avatars, such as multimedia and video games. 
For widespread adoption, capturing facial expressions from everyday devices like webcams and smartphones is crucial. 
Furthermore, for creative applications with fictional characters, the ability to \emph{retarget} captured expressions to different facial morphologies is equally important, allowing the preservation of performance expressivity while adapting to unique character appearances. 
Capturing a facial performance that can be applied to different characters necessitates the disentanglement of identity from expression information, which is our motivation for this work.

Current methods for in-the-wild facial expression capture \cite{deng2019accurate,feng2021learningDECA, danvevcek2022emoca,filntisis2023spectre,EMOTE,wang20243d, retsinas20243d} predominantly rely on 3D Morphable Models (3DMM) \cite{FLAME:SiggraphAsia2017, bfm09}, which represent facial expressions as additive vertex displacements to a neutral mesh. However, this representation has a fundamental limitation: expression-driven deformations inherently depend on identity-specific facial features, such as mouth shape or fat distribution. In 3DMMs, when the same expression parameters are applied to different identities, they produce identical vertex displacements regardless of the target's facial structure, leading to identity leakage and unnatural results. This suggests that expressions should be represented at the semantic level rather than through direct vertex displacements.



In this paper, we present SEREP (Semantic Expression Representation), a method for monocular 3D face capture that disentangles identity and expression at the semantic level. By semantic level, we mean representing expressions in a feature space that can be transformed into identity-specific vertex displacements, allowing the same underlying expression to manifest differently across various facial geometries. 
SEREP learns a semantic face expression basis from 3D expressive faces, without requiring paired expression data, FACS, or emotion labels. This expression basis allows to generate diverse synthetic faces with minimal manual effort, which we use to train our face expression capture model through a novel semi-supervised learning scheme.
By learning from both low-quality synthetic and in-the-wild data, SEREP performs accurate monocular facial expression capture, that is robust in unconstrained scenarios. 

Beyond limitations in expression representation, existing methods face significant training challenges. Some require complex self-supervised approaches with differentiable rendering for image-space reprojection \cite{deng2019accurate,feng2021learningDECA, danvevcek2022emoca,filntisis2023spectre,EMOTE,wang20243d, retsinas20243d}, while others require substantial manual effort to create synthetic training data that is as realistic as possible \cite{wood2021fake, wood20223d}. 
SEREP is trained directly in 3D space, thus circumvent the challenges associated with rendering model outputs, and generalizes to in-the-wild data, without relying on expensive and complex realistic synthetic data.


As a final contribution, we propose MultiREX, a new benchmark for facial expression capture, addressing the current lack of benchmarks for \emph{geometric} evaluation of expression capture. Current evaluation protocols typically rely on image-space metrics \cite{deng2019accurate,feng2021learningDECA,  baert2024spark, sadeghi2024unsupervised} which can introduce inaccuracies and fail to comprehensively evaluate 3D facial geometry. MultiREX, based on the Multiface dataset \cite{wuu2022multiface}, features challenging expressions with multiple viewpoints, and allows measuring geometric accuracy in 3D space.

We validate our approach through extensive experiments that show both the expressivity of our semantic expression basis and its ability to preserve identity during retargeting. SEREP outperforms state-of-the-art methods in terms of capture accuracy on MultiREX and in-the-wild data, exhibits greater robustness to viewpoint changes and better preserves identity during expression transfer.

\begin{figure*}[t]
    \centering
    \includegraphics[width=\textwidth]{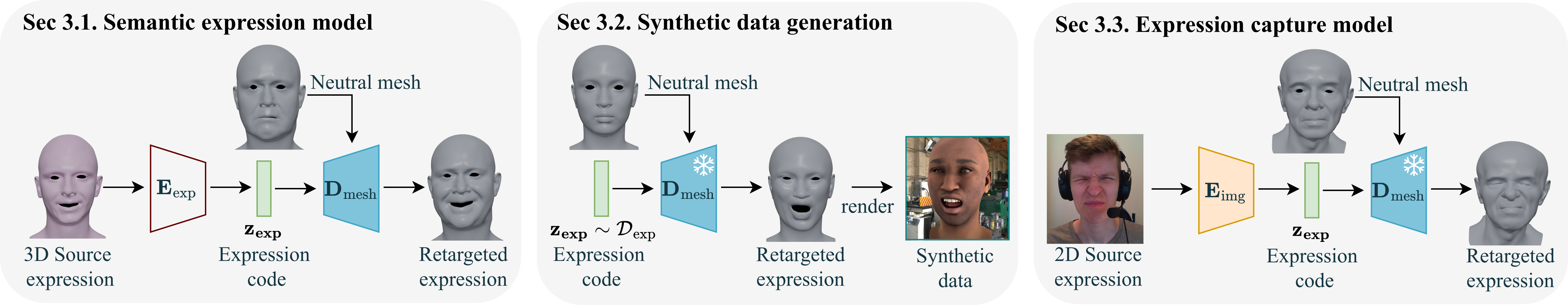}
    \caption{We first train a semantic expression model to disentangle expression from identity on 3D meshes. Next, we generate a synthetic dataset of 2D renders. Finally, we train an expression capture model on synthetic and real images, that estimates an expression code from a monocular image. This code, along with a neutral mesh, is used by $\mathbf{D}_\text{mesh}$ to obtain an expressive 3D mesh.}
    \label{fig:overview}
\end{figure*}

The contributions of our paper are as follows:

\begin{itemize}
    \item We introduce an identity-aware decoder that represents expressions at the semantic level, reducing identity leakage during expression transfer.
    \item We propose a semi-supervised training scheme using synthetic data that can be generated without substantial manual effort, while still generalizing to in-the-wild conditions.
    \item We publicly release MultiREX, the first video-based geometric 3D facial expression reconstruction benchmark.
\end{itemize}

\section{Related work}\label{sec:related_work}


\paragraph{3D Morphable Models.}
Facial performance capture from monocular video is an ill-posed problem, as the task is to recover 3D geometry from a 2D projection of the face in the image. To make this problem tractable, most recent methods \cite{feng2021learningDECA, zielonka2022towardsMICA, danvevcek2022emoca, filntisis2023spectre, EMOTE, wang20243d, lei2023hierarchical, chai2023hiface}, rely on statistical morphable models (3DMM) \cite{Egger20Years},
such as \cite{FLAME:SiggraphAsia2017, bfm09, li2020learning, chai2023hiface} to constrain the problem.
These morphable models use a large collection of scans and define: (i) a template mesh $\overline{\boldsymbol{T}}$ (average face shape), (ii) an \emph{identity} (shape) basis $\boldsymbol{S}$, and (iii) an \emph{expression} basis $\boldsymbol{E}$. An arbitrary mesh $M$ is then defined by identity ($\boldsymbol{\beta}$) and expression ($\boldsymbol{\psi}$) coefficients  of these linear bases: $M = \overline{\boldsymbol{T}} + \boldsymbol{\beta} \boldsymbol{S} + \boldsymbol{\psi}\boldsymbol{E}$. We note that a given expression vector $\boldsymbol{\psi}$ results in the same vertex displacement regardless of the identity and, therefore, cannot capture the \emph{semantic meaning} of an expression. In other words, the same semantic expression of two subjects would have different displacements and, therefore, different expression codes. 

\paragraph{Semantic expression disentanglement.}
Some recent work proposed approaches for semantic expression disentanglement. Facescape \cite{yang2020facescape} considers a corpus of multiple identities performing the same set of expressions, and represents identity and expression with a bilinear model for identity and expression \cite{vlasic2006face}. Neural Parametric Head Models (NPHM) \cite{giebenhain2023nphm} (and per extension MonoNPHM \cite{giebenhain2024mononphm}) uses a corpus of multiple identities performing the same expressions. This model operates in an implicit representation, where expression deformations are represented with an MLP that takes both expression and identity coefficients as input. We note that both methods require paired data for training (multiple subjects performing the same expression), as is commonly done in re-targeting papers \cite{chandran2022local}. This has two drawbacks: (i) it is costly to acquire, (ii) for a given expression label, they assume that the subjects are performing exactly the same facial movements. In reality, subjects are not always capable of following the exact instructions for these poses.
Other works \cite{zhao2024media2face, chandran2022facial} require facial animations to be decomposed in FACS-based rigs, a task that normally requires a significant amount of manual work.
In contrast, we learn a semantic expression space with unpaired data, without any labels or rig decomposition, in a more similar vein to the work done in face reenactment \cite{drobyshev2022megaportraits, xu2024vasa}.




\paragraph{In-the-wild face capture.}
Most recent methods for in-the-wild face capture train models in a large collection of in-the-wild images, in a self-supervised manner \cite{feng2021learningDECA, zielonka2022towardsMICA, danvevcek2022emoca, filntisis2023spectre, EMOTE, dib2024mosar, wang20243d, liu2025teaser}. These methods estimate the identity and expression parameters of a 3DMM, as well as camera, light, and reflectance information of a scene, then use a differentiable renderer \cite{kato2020differentiable, tewari2020state} to render a synthetic image of the parametrized scene. The model is trained to minimize the difference between rendered images and ground truth, together with auxiliary losses.  This camera-space optimization can sometimes result in implausible expressions when viewed from other angles, notably when the processed subject is not facing the camera. 


Another research direction shows that training models on photo-realistic synthetic data can generalize to in-the-wild capture \cite{wood2021fake, wood2022dense, hewitt2024look}. These methods render large datasets of synthetic images by controlling: identity, expression, texture, hair, clothes, and accessories. They use a large collection of artist-made and procedurally generated assets to reduce the gap between synthetic and real data.

In this work, we introduce a mixed training procedure, using real and synthetic data. Rather than attempting to close the synthetic-to-real domain gap through complex 3D modeling, we use a simple procedure to obtain synthetic data and bridge the domain gap with a domain adaptation strategy.

\paragraph{Performance evaluation.}  

Most existing work on monocular face reconstruction provides qualitative results along with either (i) geometric error in reconstructing neutral meshes or (ii) quantitative results using image-centric metrics. NOW \cite{RingNet:CVPR:2019} and REALY \cite{chai2022realy} benchmarks are commonly used to evaluate neutral mesh geometric reconstruction error, but do not account for expression recovery. Image-centric metrics include emotion classification \cite{danvevcek2022emoca,retsinas20243d}, and the Intersection-Over-Union of segmented face regions in the predicted mesh, compared to a pseudo-ground truth \cite{baert2024spark, wang20243d}.
FaceWarehouse \cite{cao2013facewarehouse} enables 3D expression evaluation instead, but only offers low-quality RGBD scans without multiview capture. In this work, we propose a new public benchmark for geometric expression capture based on the Multiface \cite{wuu2022multiface} dataset. This dataset was collected from a light stage with multiple cameras, and provides high-quality reconstructions with corresponding monocular videos under controlled camera angles.

\section{Method}


We propose a three-stage approach to capture facial expressions from in-the-wild images (\cref{fig:overview}). 
First, we train a semantic expression model (\cref{sec:method_id_exp}) that disentangles expression from identity in 3D facial meshes. This model's retargeting ability is leveraged in the second stage (\cref{sec:method_synth}) to generate 2D synthetic data to which we know the corresponding expression codes. In the third stage (\cref{sec:method_face_exp}), we train a facial performance capture model to predict expression codes from monocular images. In the absence of target expression codes for in-the-wild data (e.g., no corresponding 3D ground-truth), the model learns to regress expression codes for synthetic data and relies on a domain adaptation strategy with real data to bridge the domain gap. 




\begin{figure}
    \centering
        \includegraphics[width=1\linewidth]{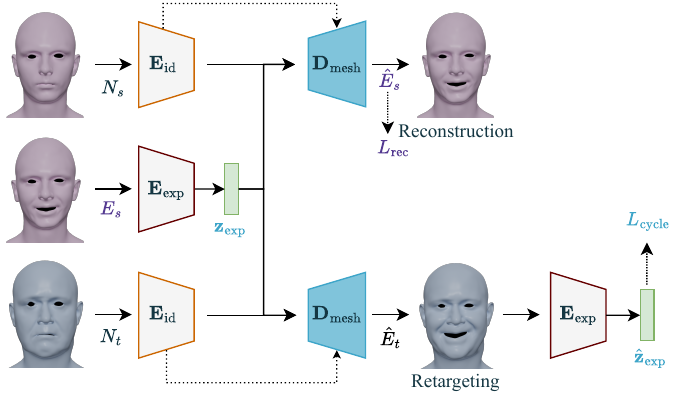}
    \caption{Semantic expression model training. The model learns to reconstruct a source expression $E_s$ given the source neutral $N_s$, and to retarget it to a target neutral $N_t$. The cycle loss aligns the codes $\textbf{z}_\text{exp}$ for the same semantic expressions of different subjects.
    }
    \label{fig:Id_exp_v5}
\end{figure}

\subsection{Semantic expression model}\label{sec:method_id_exp}

The task of the semantic expression model is to represent the expression from an expressive mesh $E_s$ and apply it to a neutral mesh of the same source subject $N_s$ (i.e., reconstruction) or a different target subject $N_t$ (i.e., retargeting). For training, the model learns those two objectives together, as illustrated in \cref{fig:Id_exp_v5}.




This model is composed of an encoder $\mathbf{E}_\text{id}$ that encodes identity, an encoder $\mathbf{E}_{\text{exp}}$ that encodes expression, and a single decoder $\mathbf{D}_\text{mesh}$ that decodes the final mesh. All models operate directly on mesh vertices, using spiral convolutions \cite{bouritsas2019neural, COMA:ECCV18}. 
We describe the loss terms used in training below, where the first two are designed for learning a semantic expression space, while the remaining serve as regularization losses.

\noindent \textbf{Reconstruction loss.} 
We use the squared $L2$ distance between vertices of the source expressive mesh $E_s$ and its reconstruction $\hat{E}_s$: $L_\text{rec} = \lVert \hat{E}_s - E_s \rVert^2$.


\noindent \textbf{Cycle consistency loss.} 
The model should represent the same semantic expressions for different subjects with the same expression code. Without paired expression data, there is no ground truth for the retargeted mesh. 
Thus, we use a cycle consistency loss \cite{ZhouCycle} to encourage semantic expressions to remain the same after retargeting on the target identity: ${L}_\text{cycle} = \lVert \mathbf{z}_\text{exp} - \mathbf{E}_\text{exp}(\hat{E}_t) \rVert^2 $.

\noindent \textbf{Edge preservation loss.} 
We use edge length regularization $L_\text{edge}$, similar to \cite{Bolkart2023Tempeh}, to reduce jagged edges on the decoded meshes. This loss penalizes large changes in edge length between the retargeted mesh $\hat{E}_t$ and the neutral $N_t$.

\noindent \textbf{Eye closure loss.} 
This term, noted $L_\text{eyes}$ enforces eye closure. We compute the distance between the two polylines defined by the sequence of vertices of the upper and the lower eyelid. This term is enforced when the source expression $E_s$ contains a closed eye.



\noindent \textbf{Delta loss.} We complement our data set of expressive meshes with an additional set of neutral meshes to improve generalization to new identities. These additional target neutral meshes $N_t$ require a supplementary loss term to stabilize the training. This term, defined as $L_\text{delta} = \lVert (\hat{E}_t - N_t) - (E_s - N_s) \rVert^2 $ prevents the decoder from collapsing to a trivial solution and producing neutral meshes regardless of the expression code.

The complete training loss for the retargeting model $\mathcal{L}_\text{ret}$ is consequently defined as follows: 
\begin{equation}
    \mathcal{L}_\text{ret} =  L_\text{rec} + L_\text{cycle} + L_\text{edge} + L_\text{eyes} + L_\text{delta}
    \label{eq:retarget}
\end{equation}
Importance weights are omitted for better readability and detailed in supplementary material. 


\begin{figure}
        \centering
       \includegraphics[width=1\linewidth]{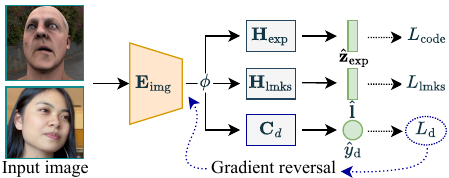}
        \caption{Expression capture model training. Given a synthetic or real input image, the model encodes features $\phi$ and estimates the expression code $\textbf{z}_\text{exp}$, landmarks $\mathbf{l}$, and the domain (real or synthetic). Gradient reversal and the classifier $\mathbf{C}_d$ are used to bring real and synthetic images close in feature space.
        }
        \label{fig:Face_capture_v3}
    \end{figure}

\subsection{Synthetic data generation}\label{sec:method_synth}

Once the semantic expression model is trained, we exploit it to generate a synthetic dataset that will be used to train the face expression capture model. First, we sample a target identity $N_t$ and expression code $\textbf{z}_\text{exp}$ from the dataset and produce an expressive mesh $\hat{E}_t$. We procedurally place a set of teeth and eyes, and a random face albedo texture, environment map, and head pose. Finally, we render a synthetic image using \cite{blender2021cycles}. Unlike \cite{wood2021fake}, we do not add accessories, hair, or clothes to the scene, avoiding any further 3D modeling. Examples of the generated images can be visualized in the supplementary material. We also extract 2D facial landmarks, denoted $\mathbf{l}$, by projecting the corresponding mesh vertices to camera coordinates. 

\subsection{Expression capture model}\label{sec:method_face_exp}
Our expression capture model estimates semantic expression codes from images, using a joint training strategy with synthetic and real data (\cref{fig:Face_capture_v3}). The encoder $\mathbf{E}_{\text{img}}$ extracts latent features $\phi$  which feed three task-specific components: the $\mathbf{H}_{\text{exp}}$ head estimates the semantic expression code $\hat{\textbf{z}}_\text{exp}$ using the ground truth codes from the synthetic data, $\mathbf{H}_{\text{lmks}}$ estimates 2D landmarks $\hat{\mathbf{l}}$, and $\mathbf{C}_d$ classifies images as real or sythetic. The last two are used to align features across domains to improve generalization to in-the-wild images.



The encoder $\mathbf{E}_{\text{img}}$ uses a ConvNeXt-B backbone \cite{liu2022convnet}. The $\mathbf{H}_{\text{code}}$ head and the domain classifier $\mathbf{C}_d$ each consists of 3 blocks of linear layers and group normalization \cite{wu2018group}, with skip-connections \cite{he2016identity}. The $\mathbf{H}_{\text{lmks}}$ head is a single linear layer. 
The low capacity of the $\mathbf{H}_{\text{lmks}}$ head encourages the encoder to produce features $\phi$ that encode the necessary information for landmark prediction, the task shared between both domains.

Our training loss comprises three terms:

\noindent \textbf{Latent code loss.} 
This term encourages an accurate prediction of the expression code, but can only be applied to synthetic samples since we do not have ground truth expression data for in-the-wild images. It is defined as $L_\text{code} = \lVert \hat{\textbf{z}}_\text{exp} - \textbf{z}_\text{exp} \rVert^2$.



\noindent \textbf{Landmarks loss.} This term is the pixel distance between the predicted landmarks and the GT landmarks defined as: $L_\text{lmks} = \lVert \hat{\mathbf{l}} - \mathbf{l} \rVert^2$. 
Landmark prediction is the shared task between both domains; therefore, the loss term applies to all images, real and synthetic. It ensures that the features $\phi$ encode information about salient regions of the face regardless of the image domain.

\noindent \textbf{Domain loss.} To minimize the domain gap between synthetic and real data, we adopt the domain adaptation approach from \cite{ganin2015unsupervisedgradientreversal}, and define the corresponding domain loss $L_d$. The classifier $\mathbf{C}_d$ is trained to distinguish between real and synthetic images. During gradient descent optimization, we reverse the sign of the gradient after $\mathbf{C}_d$ and scale its magnitude by a factor $\lambda$. As a result, $\mathbf{C}_d$ is optimized to discriminate between the two domains, while the features $\phi$ are encouraged to discard domain-specific information, thus aligning the distributions of both domains more closely. 



The training loss for this model is defined as follows:
\begin{equation}
                \mathcal{L}_\text{cap} =  L_\text{code} + L_\text{lmks} + L_\text{d}
\label{eq:capture}
\end{equation}
Importance weights in \cref{eq:capture} are omitted for clarity and detailed in the supplementary material. 

\section{Experiments on expression representation}


\subsection{Implementation detail}
We train the expression model using two datasets: static neutral meshes, and expressive sequences. We use neutral scans from 865 individuals, and dynamic sequences from 10 subjects, each captured over sessions lasting 3 to 19 minutes. These sequences cover a variety of expressions and talking sequences, which should reasonably cover the expression space naturally bounded by human physiology. We evaluate our results on static scans of 56 subjects, including a neutral mesh and 19 fixed expressions. All meshes share the same topology, with 13k vertices. 

We jointly train the encoders $\mathbf{E}_\text{id}$, $\mathbf{E}_\text{exp}$, and the mesh decoder $\mathbf{D}_\text{mesh}$, to minimize \cref{eq:retarget} using the Adam  optimizer \cite{kingma2014adam}. Training on a single A4000 GPU took 3 days.

In all experiments, we use the official implementation of each baseline
For quantitative experiments, results reported in bold are statistically significant based on a Wilcoxon signed rank test ($\rho<0.001$).

\subsection{Expression reconstruction}
\label{ssec:exp_recon}
First, we evaluate the ability of the proposed method to reconstruct expressive face meshes for the 56 identities of the test set (i.e., evaluating for 1064 meshes). For each expressive mesh, we define an optimization problem: given the subject's neutral and expressive meshes, find the expression coefficients that minimize the reconstruction loss.  We compare our method to the widely used FLAME model \cite{FLAME:SiggraphAsia2017}. 

In \cref{tbl:comparison}, we report the mean reconstruction error, averaging over all meshes, and considering only a frontal mask of the face. Our method outperforms FLAME, indicating that our expression model is better able to capture the geometric deformations caused by expressions on unseen subjects. We hypothesize this improvement stems in part because it is a non-linear model, which is often more expressive than their linear counterparts. Interestingly, the generalization capability does not come at the price of larger training corpus given that our model was trained using neutral meshes from 865 subjects and animated meshes from 10 subjects, while the FLAME model was trained using 3800 neutral subjects and animations from 24 subjects.

\subsection{Identity preservation}
For this evaluation, we consider all pairs of [source, target] subjects from the test set. For each expressive mesh, we optimize the expression code (as in \cref{ssec:exp_recon}) and apply it to the target's neutral mesh. In total, we obtain 61,600 retargeted meshes for each method. We render the retargeted meshes using the GT neutral texture and a 50mm camera using Blender. Finally, we measure the similarity between the rendered images of the GT and the retargeted images using the cosine similarity (CSIM) of identity embedding as commonly done in face re-enactment research \cite{ren2023hr,drobyshev2022megaportraits,hsu2022dual}.
A higher CSIM indicates a better preservation of the identity. In other words, a lower source identity traits leakage on the retargeted mesh.

\cref{tbl:comparison} reports the result of this experiment. As a lower bound, we also report the mean CSIM between random subjects with the same expression. Our method produces retargeted meshes that better preserve the identity of the target person when compared to FLAME. 
This confirms that the identity and expression bases from FLAME do not allow for fully capturing the semantics behind a given expression and that identity leakage occurs. In contrast, our semantic expression model is better able to retarget an expression preserving the target person's identity.

\subsection{Ablations on the semantic expression model}

\begin{table}
    \centering
    \resizebox{\columnwidth}{!}{ 
    \begin{tabular}{l|c|c}
        \hline
        \textbf{Method} & \textbf{Rec. Error} (\textdownarrow)& \textbf{CSIM} (\textuparrow) \\ \hline
        Random & - & 0.501 $\pm$ 0.11  \\
        FLAME & 1.36 $\pm$ 0.34 & 0.766 $\pm$ 0.08 \\
        \midrule
        Ours (No delta reg.) & 1.75 $\pm$ 0.40 & 0.736 $\pm$ 0.09 \\ 
        Ours (No ID in decoder) & \textbf{1.15 $\pm$ 0.27} & 0.773 $\pm$ 0.09 \\ 
        Ours & \textbf{1.15 $\pm$ 0.24} & \textbf{0.791 $\pm$ 0.08} \\ \hline
    \end{tabular}
    }
    \caption{Mean reconstruction error in mm and cosine identity similarity (CSIM) of retargeted expressions against ground truth.}
    \label{tbl:comparison}
\end{table}

\begin{figure}
    \centering
    \vspace{-1em}
    \includegraphics[width=1\linewidth]{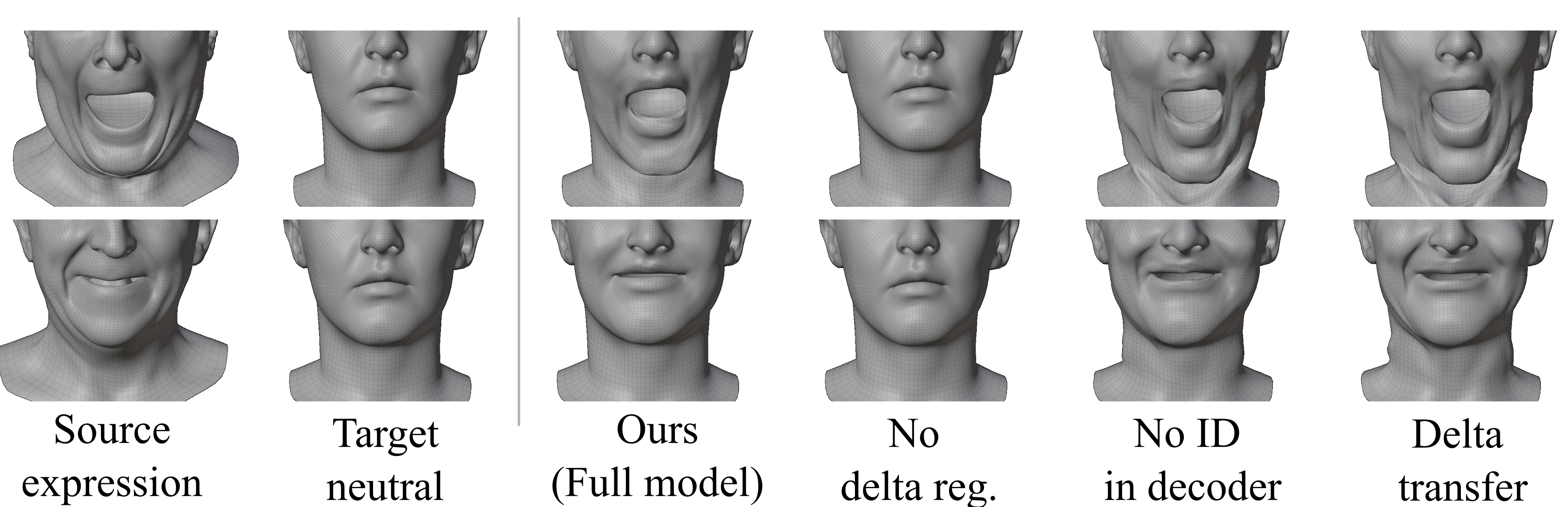}
    \caption{Ablations on the semantic expression model.}
    \label{fig:ablation_semantic_model}
\end{figure}

We investigate removing either the delta regularization term (no delta reg.) or the identity conditioning from the decoder (no ID in decoder) to validate its impact on identity-expression separation. In the latter case, the model learns to generate vertex displacements solely from expression codes that are directly added to the neutral mesh, as commonly done in 3DMMs. We also investigate removing the edge preservation and eye closure loss in supplementary material. Those two mostly enforce visual quality.  
As shown in \cref{fig:ablation_semantic_model}, without delta loss, the decoder collapses to mostly generating neutral faces, severely limiting expression reconstruction accuracy and causing significant identity leakage during retargeting (\cref{tbl:comparison}). While the \emph{no ID in decoder} model achieves comparable reconstruction error,  it learns additive displacements similar to delta transfer (\cref{fig:ablation_semantic_model}). This compromise identity preservation in retargeting, and highlights the need for identity conditioning. 

\begin{table*}
\centering
\fontsize{10}{12}\selectfont
\setlength\tabcolsep{5pt}
\resizebox{0.72\linewidth}{!}{%
\begin{tabular}{l|cccc|c}
\toprule
\textbf{Method} & \textbf{Cheek} & \textbf{Forehead} & \textbf{Mouth} & \textbf{Nose } & \textbf{Average} \\
\midrule
DECA \cite{DECA:Siggraph2021} & $3.54 \pm 1.59$ & $1.82 \pm 0.70$ & $3.71 \pm 1.81$ & $1.30 \pm 0.47$ & $2.59 $ \\
EMICA \cite{EMOTE} & $3.26 \pm 1.57$ & $\mathbf{1.65 \pm 0.72}$ & $3.44 \pm 1.86$ & $1.10 \pm 0.47$ & $2.36 $\\
SMIRK \cite{retsinas20243d} & $4.13 \pm 1.59$  & $1.84 \pm 0.66$ & $3.69 \pm 1.47$ & $1.29 \pm 0.43$ & $2.74$\\
\bottomrule
Ours (synth only) &  $2.89 \pm 1.38$ & $2.13 \pm 0.78$ & $3.05 \pm 1.67$ & $1.11 \pm 0.44$ & $2.30$ \\
Ours (no $L_\text{d}$) & $\mathbf{2.80 \pm  1.33}$ & $\mathbf{1.64 \pm  0.66}$ & $\mathbf{2.95 \pm  1.59}$ &$\mathbf{1.00 \pm  0.42}$ & $2.10$ \\
Ours &  $2.83 \pm 1.28$  & $1.89 \pm 0.81$ & $\mathbf{2.95 \pm 1.51}$ & $1.08 \pm 0.45$ & $2.19$ \\
\bottomrule
\end{tabular}
}
\caption{Expression reconstruction results on  MultiREX. We show per-vertex average errors (in mm) on different facial regions.}
\label{tab:region_wise_ave}
\end{table*}

\section{Experiments on capture and retargeting}
In this section, we evaluate the performance of SEREP (\cref{sec:method_face_exp}) on two applications: expression capture and retargeting.  
We first introduce MultiREX, our proposed open-source benchmark, which enables comparison of geometric expression capture.


\subsection{MultiREX Benchmark}\label{sec:multirex_benchmark}

MultiREX (Multiface Region-based Expression evaluation) is based on the Multiface dataset \cite{wuu2022multiface}. It evaluates the estimated geometry of monocular face capture systems considering complex expression sequences under multiple camera views. In particular, the protocol evaluates mesh deformations related to expression alone, treating the identity as a given.

The benchmark includes 8 identities captured simultaneously from five viewpoints: \emph{Frontal}, two \emph{Angled} views (yaw rotation around 40 degrees), and two \emph{Profile} views (yaw rotation around 60 degrees).
Each subject performs a range-of-motion sequence covering a wide range of expressions, including extreme and asymmetrical motions. The benchmark comprises 10k ground truth meshes and 49k images. 

We obtain the ground truth identity (i.e., neutral mesh) by manually selecting a neutral frame for each subject and retopologizing the corresponding mesh to the FLAME topology using commercial software. From these two meshes, we compute a per-subject sparse conversion matrix that enables fast conversion from the FLAME to the Multiface topology.


Inspired by the REALY benchmark \cite{chai2022realy}, we adopt a region-based evaluation method, dividing the face into four regions (i.e., forehead, cheek, mouth, and nose) and performing region-based rigid alignment before assessment. This avoids penalizing a model due to rigid misalignment between the predicted and GT meshes, and instead focuses on the non-rigid deformations. For each region, we find the optimal rigid alignment between the GT and predicted meshes in the Multiface topology and compute the per-vertex error. Masks and alignment details are provided in the supplementary material.

We publicly release\footnote{Link to code and assets: \url{https://github.com/ubisoft/ubisoft-laforge-multirex}}: (i) the code to download assets, (ii) neutral meshes in the FLAME topology alongside the code to convert between FLAME and Multiface topologies, (iii) code to run the benchmark and compute the metrics.

\subsection{Implementation details of the capture model}
We train the face expression capture model using in-the-wild and synthetic images. For in-the-wild data, we use WFLW \cite{wu2018look}, which contains 10k face images annotated with landmarks. We complement with CelebV-HQ \cite{zhu2022celebvhq} from which we sample four frames per video and apply an off-the-shelf landmark detector \cite{zhou2023star} to obtain pseudo ground truth. This amounts to 165k images.

In addition, we generate 135k synthetic images with corresponding codes, following the procedure described in \cref{sec:method_synth}. We also project mesh vertices to camera coordinates to obtain ground truth landmarks in image space, except for eyebrows. Eyebrow landmarks are not precise for mesh vertices \cite{Ferman_2024_CVPR} as they also depend on texture for synthetic data, so we use a landmark detector to obtain them \cite{zhou2023star}.

For the following experiments, given an identity mesh $N$ and a source image $X$, we obtain the expressive mesh with $M = \mathbf{D}_\text{mesh}(\mathbf{E}_\text{exp}(X), \mathbf{E}_\text{id}(N))$. For the other methods, we use their image encoders to obtain the expression coefficients $\boldsymbol{\Psi}$, and apply them directly to the neutral mesh with $M = N + \boldsymbol{\Psi}\boldsymbol{E}$. For FLAME-based models, this process is followed by the Linear Blend Skinning (LBS) function to account for jaw rotation. For experiments on ITW data neutrals are obtained with an off-the-shelf reconstruction method \cite{dib2024mosar}.

During training, we minimize \cref{eq:capture} using the Adam optimizer \cite{kingma2014adam}. We apply standard data augmentations such as translations, rotations, scale, and occlusions with random patches. Training took 11 hours on two A4000 GPUs.



\subsection{Results on MultiREX} 
\label{sec:multirex_results}
We now show quantitative results on geometric expression capture, on the proposed MultiREX benchmark. We process videos from all subjects and camera views, comparing our model to DECA \cite{DECA:Siggraph2021}, EMICA \cite{danvevcek2022emoca, EMOTE} and SMIRK \cite{retsinas20243d}. The neutral mesh is manually selected (\cref{sec:multirex_benchmark}), and considered as a given for all approaches. 
Therefore MultiREX evaluation only accounts for geometric deformations due to the facial expressions. Additionally, we use ground truth bounding boxes, computed from projecting keypoint vertices to the image space. Results reported in bold are statistically significant based on a Wilcoxon signed rank test ($\rho<0.001$).




\begin{table}
\centering
\fontsize{10}{12}\selectfont
\setlength\tabcolsep{5pt}
\resizebox{\columnwidth}{!}{%
\begin{tabular}{l|ccc}
\toprule
\textbf{Method} & \textbf{Frontal} & \textbf{Angled} & \textbf{Profile} \\
\midrule
DECA \cite{DECA:Siggraph2021} & $2.57 \pm 1.58$ & $2.61 \pm 1.66$ & $2.61 \pm 1.70$ \\
EMICA \cite{EMOTE} & $2.17 \pm 1.40$ & $2.30 \pm 1.55$ & $2.49 \pm 1.77$ \\
SMIRK \cite{retsinas20243d} & $2.25 \pm 1.25$ & $2.58 \pm 1.61$  & $3.11 \pm 1.80$ \\
\midrule
Ours & $\mathbf{2.08 \pm 1.19}$ & $\mathbf{2.08 \pm 1.29}$ & $\mathbf{2.30 \pm 1.40}$ \\

\bottomrule
\end{tabular}
}
\caption{MultiREX expression reconstruction results on frontal and side views. }\label{tab:view_wise_performance}
\end{table}

\cref{tab:region_wise_ave} reports the average reconstruction error in millimeters for different regions of the face and the average over all face regions. 
Our method outperforms other state-of-the-art methods on average and on all individual regions except for the forehead region. These improvements over competing methods are explained by the expressivity of our representation and the robustness of SEREP to viewpoints. 

\begin{figure}
    \centering
    \includegraphics[width=1\linewidth]{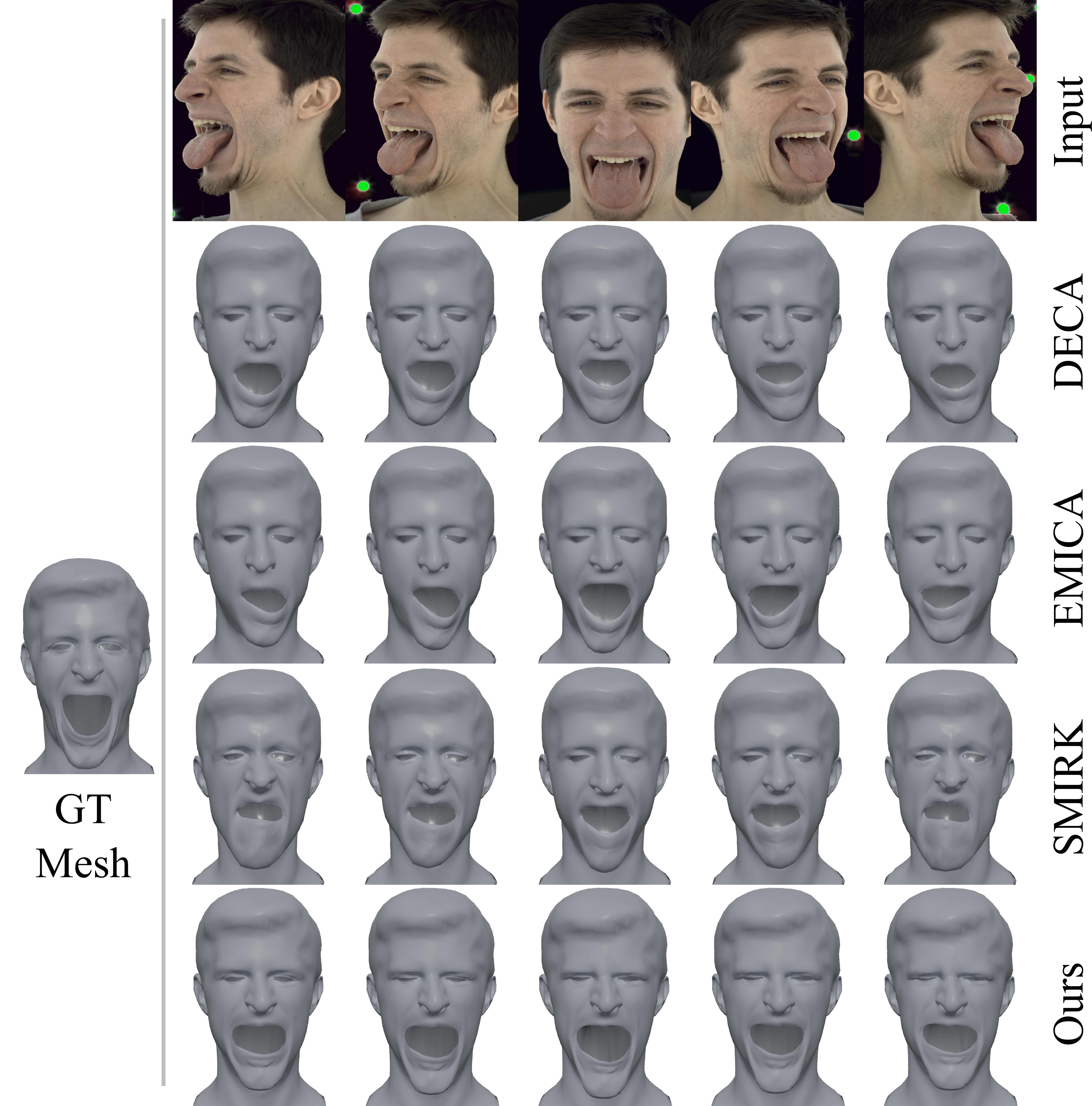}
    \caption{Facial expression capture examples on  MultiREX. Left: Ground truth (GT) mesh of the target subject. Right: predictions for different models on each camera view.}
    \label{fig:qualitative_multirex_multiviews}
\end{figure}

We further analyze the viewpoint robustness of each method by reporting the average reconstruction errors for each viewing angle independently in \cref{tab:view_wise_performance}.  These results indicate that our model maintains performance better than competing methods as the viewing angle shifts from the frontal view. 
In contrast, the performance of EMICA and SMIRK decreases faster as the camera angle increases. 
This can be visualized in \cref{fig:qualitative_multirex_multiviews}, where we show results for the same expression captured from five different views. 
Additional examples are provided in the supplementary material. 
Our model produces consistent results across views, particularly in side views when compared to competing methods. This is especially apparent in the mouth shape of the subject. 
This is explained by the fact that our method 
relies on a semantic space that represents facial expressions in their entirety, 
rather than relying solely on image space information where some regions may be occluded.

\subsection{In-the-wild retargeting results}


\begin{figure}
    \centering
    \includegraphics[width=1\linewidth]{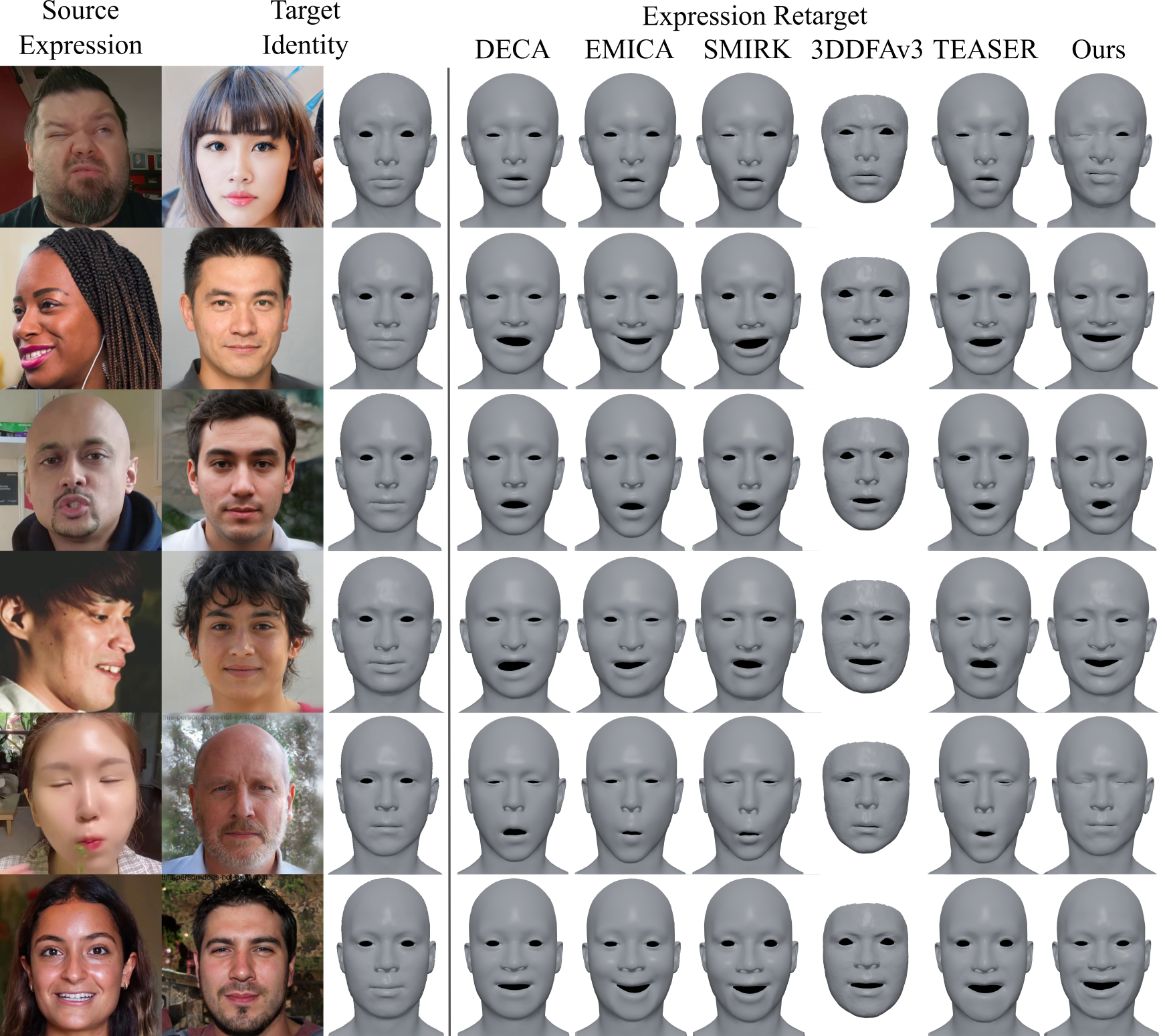}
    \caption{In-the-wild expression retargeting results. Expressions estimated on the source image are applied to a target identity. }
    \label{fig:inthewild_v7}
\end{figure}

In this section, we compare our method with recent state-of-the-art techniques, DECA \cite{feng2021learningDECA}, EMICA \cite{EMOTE}, SMIRK \cite{retsinas20243d}, 3DDFA-V3 \cite{wang20243d}, and TEASER \cite{liu2025teaser} on the expression retargeting task from in-the-wild images. Since ground truth data for corresponding expressions across subjects on in-the-wild images is not possible to obtain, we compare methods qualitatively. We show several examples in \cref{fig:inthewild_v7}, where the expressions from the images on the first column are transferred to the target identities of the second one.

As can be observed in these examples\footnote{More examples in the supplementary material}, SEREP reproduces challenging asymmetrical expressions better than the other methods. For instance, it completely shuts the eyes and closes the mouth on lines 1 and 5. On line 3, SEREP produces a funnel mouth. On the side views of the second and fourth lines, SEREP produces more plausible facial expressions than other methods, which corroborate the results of \cref{sec:multirex_results}. This illustrates the benefits of representing expressions at the semantic level instead of in terms of vertex displacements. A wink, a crooked smile, and a mouth pucker semantically mean the same for everyone and do not depend on viewpoint. 
Lastly, while our model is trained on individual frames, we obtain time-consistent results applying it to videos. The supplementary video showcases this result for both MultiREX and in-the-wild videos.

\subsection{Ablations on the expression capture model}

\begin{figure}
        \centering
        \includegraphics[width=1\linewidth]{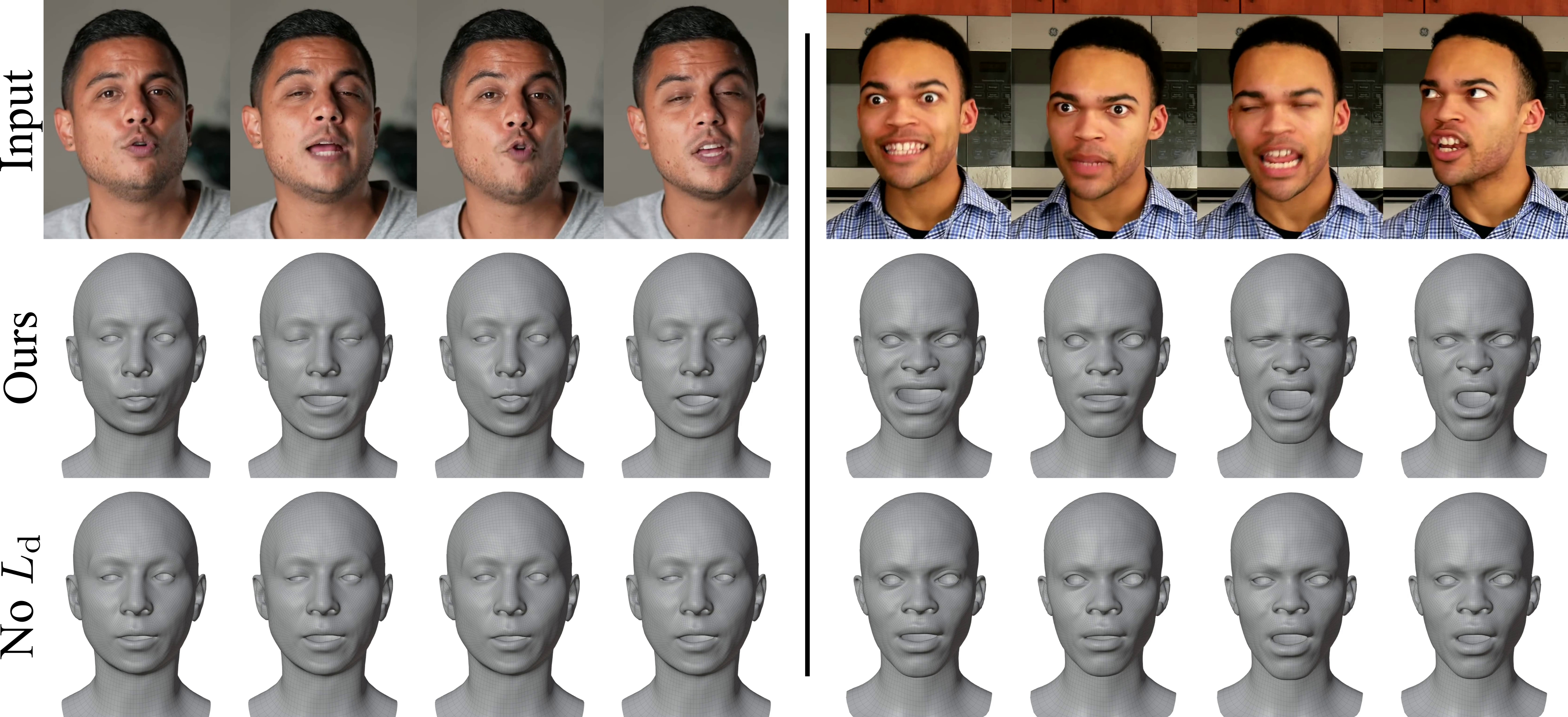}
        \caption{Impact of the domain loss $L_\text{d}$ on in-the-wild capture.}
        \label{fig:domain_loss_ablation}
        \vspace{-1.3em}
    \end{figure}

We conduct ablation studies to evaluate two key aspects of our approach: (i) the necessity of combining synthetic and real data, and (ii) the impact of the domain loss $L_d$. 
In \cref{tab:region_wise_ave}, we see that training only on synthetic data leads to lower performance, indicating that mixing the synthetic and real domains is beneficial. 
Interestingly, still in \cref{tab:region_wise_ave}, we see that the domain loss ($L_d$) slightly degrades performance on MultiREX. However, it is necessary for proper generalization to in-the-wild domain, as observed in \cref{fig:domain_loss_ablation}. Results without $L_d$ exhibit limited motion and expressiveness, which is even more evident when viewing dynamic sequences in our supplementary video. $L_d$ introduces a trade-off between performance in a controlled lightstage environment and generalization to in-the-wild data. 
This trade-off is difficult to quantify, given that benchmarks for geometry accuracy like MultiREX do not exist in unconstrained settings. 



\section{Limitations}


We focus on facial expression capture and consequently assume a known identity, therefore not estimating the neutral mesh. We acknowledge that the methods used for comparison in our experiments, while being the best candidates to our knowledge, were designed to address the broader task of reconstructing facial shape, expression and textures.

In addition, our benchmark currently only supports the FLAME topology. Although FLAME is the most widely used 3DMM, extending the benchmark to include other topologies such as Basel \cite{bfm09}, ICT \cite{li2020learning}, and HiFace \cite{chai2023hiface}, would facilitate the comparison of a broader range of methods.
Finally, our method struggles with external occluders such as glasses and inflated cheeks given that they are not represented enough in the training data. 
\section{Conclusion}


In this work, we present SEREP, an expression capture model that operates at the semantic level, enabling precise capture and retargeting of facial expressions. We start by learning a semantic expression model to represent expressions from unpaired data and directly in the 3D space. This model is leveraged to generate synthetic data with minimal manual labour, after sampling and retargeting known expression codes. Finally, we introduce a semi-supervised learning scheme that enables our expression capture model training, using the synthetic data and a domain adaptation strategy to generalize to unconstrained conditions.   

Our experiments underscore the strengths of our expression representation, demonstrating its effectiveness in capturing expressions under various conditions, including challenging viewpoints and luminosity, while accounting for the identity facial geometry. Additionally, we introduce MultiREX, a new benchmark that evaluates expression geometry directly in 3D space. For future work, we aim to make our capture model temporally aware to improve video capture quality and plan to integrate camera position information for enhanced supervision.


\section{Acknowledgements}

This research was supported by the \href{https://www.nserc-crsng.gc.ca/index_eng.asp}{Natural Sciences and Engineering Research Council of Canada (NSERC)} and \href{https://www.mitacs.ca/}{MITACS} through the \textit{Alliance-Mitacs} program (Grant No. ALLRP 589317 - 23).

We would also like to thank authors of the Multiface dataset \cite{wuu2022multiface}, as well as authors of DECA \cite{DECA:Siggraph2021}, EMICA \cite{EMOTE}, SMIRK \cite{retsinas20243d}, 3DDFAv3 \cite{wang20243d}, and TEASER \cite{liu2025teaser} for releasing their code and advancing research in the field.



{\small
\bibliographystyle{ieee_fullname}
\bibliography{egbib}
}

\appendix

\section*{Appendices} 
\section{Implementation details}
For training the semantic expression model, we use the following importance weight in eq. 1 of the main manuscript. 
$\lambda_{rec} =  1.0$, $\lambda_{cycle} =  1.0$, $\lambda_{delta} =  0.01$, $\lambda_{edge} =  10000$ and $\lambda_{eyes} =  0.01$. Details for the model's architecture are provided \cref{tab:semantic_model_desc}.

For the expression capture model, the following importance weight are used in eq. 2 of the main manuscript: $\lambda_{code} =  10$, $\lambda_{lmks} =  1$ and $\lambda_{domain} =  0.005$. For the gradient reversal, we use a scale factor equal to 1. Details for the model's architecture are provided \cref{tab:face_capture_desc}.

\begin{table*}[h!]
    \centering
    \begin{tabular}{|l|p{12cm}|}
        \hline
        \textbf{Function} & \textbf{Details} \\ \hline
        $\mathbf{E}_\text{exp}$ & 
        \begin{itemize} 
            \item \textbf{SpiralConv (x5)} \texttt{(3, 32)} $\to$ \texttt{(32, 32)} $\to$ \texttt{(32, 32)} $\to$ \texttt{(32, 64)} $\to$ \texttt{(64, 64)}
            \item \textbf{Linear:} \texttt{Linear(3392, 64)}
        \end{itemize} \\ \hline
        
        $\mathbf{E}_\text{id}$ & 
        \begin{itemize}
            \item \textbf{SpiralConv (x5)} \texttt{(3, 32)} $\to$ \texttt{(32, 32)} $\to$ \texttt{(32, 32)} $\to$ \texttt{(32, 64)} $\to$ \texttt{(64, 64)}
            \item \textbf{Linear:} \texttt{Linear(3392, 64)}
        \end{itemize} \\ \hline
        
        $\mathbf{D}_\text{mesh}$ & 
        \begin{itemize}
            \item \textbf{Linear:} \texttt{Linear(128, 3392)}
           \item \textbf{SpiralConv (x5)} \texttt{(64, 64)} $\to$ \texttt{(64, 32)} $\to$ \texttt{(32, 32)} $\to$ \texttt{(32, 32)} $\to$ \texttt{(32, 3)}
        \end{itemize} \\ \hline

    \end{tabular}
    \caption{Model architecture for the semantic expression model. }
    \label{tab:semantic_model_desc}
\end{table*}

\begin{table*}[h!]
    \centering
    \begin{tabular}{|l|p{12cm}|}
        \hline
        \textbf{Function} & \textbf{Details} \\ \hline
        $\mathbf{E}_\text{img}$ & \begin{itemize} \item \textbf{ConvNeXt-B()} \end{itemize}
        \\ \hline
        
        $\mathbf{H}_\text{code}$ & 
        \begin{itemize}
            \item \textbf{ResBlock (x3):} \texttt{Linear(512, 512)} $\to$ \texttt{GELU} $\to$ \texttt{GroupNorm(32, 512)}
            \item \textbf{Linear:} \texttt{Linear(512, 64)}
        \end{itemize} \\ \hline
        $\mathbf{H}_\text{lmks}$ & 
        \begin{itemize}
            \item \textbf{Linear:} \texttt{Linear(512, 128)}
        \end{itemize} \\ \hline
        
        $\mathbf{C}_\text{d}$ & 
        \begin{itemize}
            \item \textbf{GradientReversal()}
            \item \textbf{Linear:} \texttt{Linear(512, 256)} $\to$ \texttt{GroupNorm(16, 256)} $\to$ \texttt{GELU}
            \item \textbf{ResBlock (x2):} \texttt{Linear(256, 256)} $\to$ \texttt{GELU} $\to$ \texttt{GroupNorm(16, 256)}
            \item \textbf{Linear:} \texttt{Linear(256, 1)}
        \end{itemize} \\ \hline
    \end{tabular}
    \caption{Model architecture for the expression capture model.}
    \label{tab:face_capture_desc}
\end{table*}

\section{MultiREX additional details}
We built the MultiREX benchmark from 8 identities from the MultiFace dataset \cite{wuu2022multiface}.
%
We selected a single Range-of-Motion (ROM) sequence per identity, where the subjects perform a large variety of facial movements, including extreme expressions.
For each ROM video, we consider 5 camera views, with the exception of identity `002914589' which only includes 4 (due to a camera failure). In total, we use 39 distinct videos. The benchmark comprises 10k ground truth meshes and 49k images. 
We note that while the original Multiface dataset contains 13 identities, we did not consider 5 subjects that either: (i) did not contain the range-of-motion sequence, (ii) had a camera failure for the frontal video or (iii) had videos that cropped a large portion of the subject's face.

\begin{figure}
    \centering
    \includegraphics[width=1\linewidth]{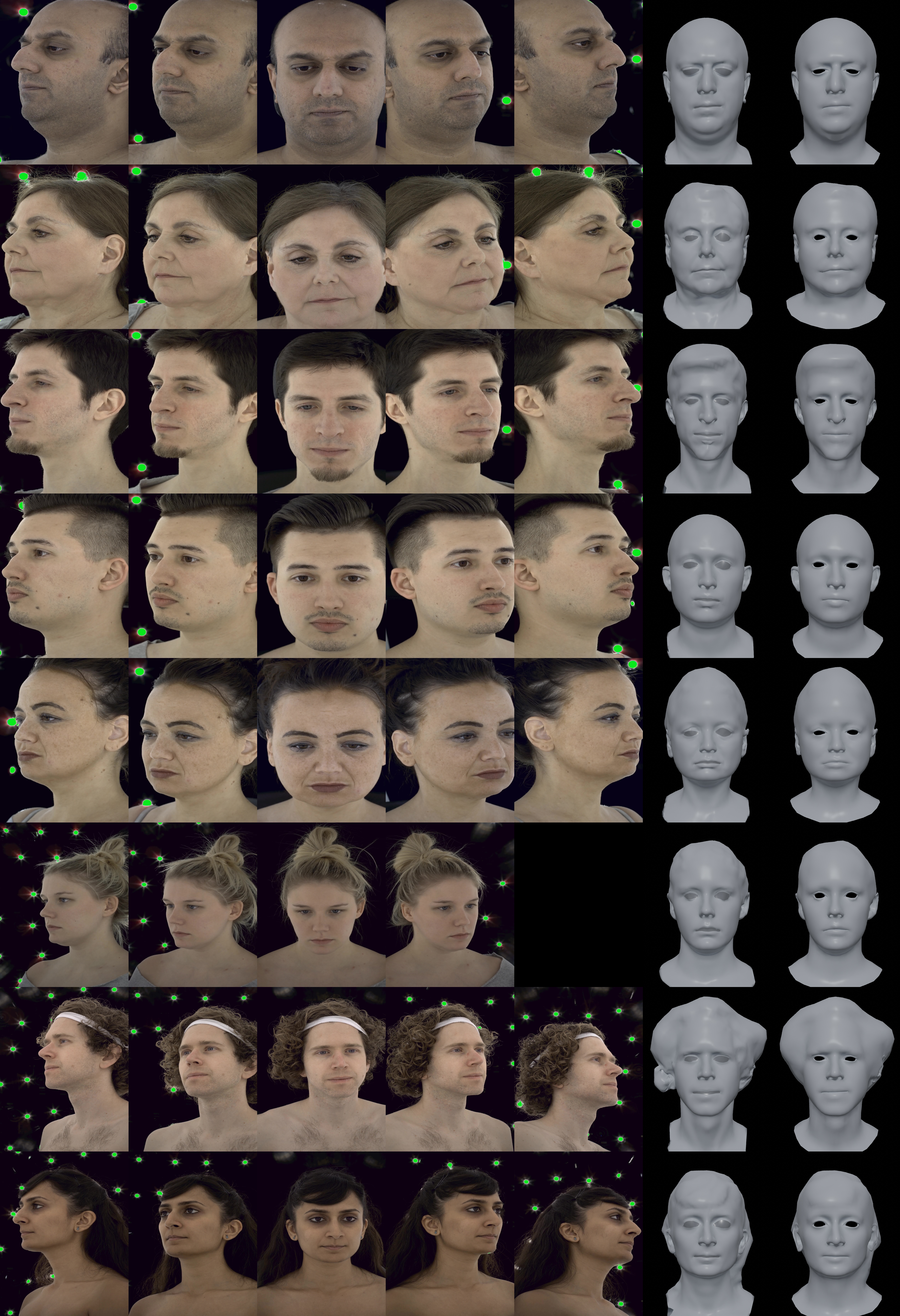}
    \caption{The 5 camera views used in MultiREX (left-to-right), followed by the corresponding ground-truth mesh under the Multiface and FLAME topology. We show all 8 subjects in the benchmark.}
    \label{fig:MultiREX_benchmark}
\end{figure}

\cref{fig:MultiREX_benchmark} presents a frame for the different views used for evaluation, using the frame we manually selected for neutral representation of each individual. This is followed by the corresponding ground-truth mesh under the multiface topology and finally by the wrapped equivalent under the FLAME topology. FLAME neutrals are obtained using commercial software (Wrap 3D\footnote{\url{https://faceform.com/}}), by first aligning each multiface mesh to the FLAME basehead with a rigid alignment with manually selected keypoints around eyes, nose, and mouth, then wrapping the mesh for topology conversion.

\section{Ablations on the semantic expression model}

We ablate the eye closure loss $L_\text{eyes}$ and edge loss $L_\text{edge}$ in \cref{fig:ablation}. Without $L_\text{eyes}$, eye closure is incomplete during blinks. 
The edge loss follows existing practice \cite{Bolkart2023Tempeh} and improves animations' edge flow (removing jagged lines) to better support downstream animator needs. We will include these figures in the supplementary material.

\begin{figure*}
    
    \centering
    \includegraphics[width=\linewidth]{figures/ablations_iccv.pdf}
    \vspace{-1.5em}
      \caption{Ablation on $L_\text{eyes}$ and $L_\text{edge}$.}
    \label{fig:ablation}
    \vspace{-1.8em}
\end{figure*}

\begin{figure}
    \centering
    \includegraphics[width=0.6\linewidth]{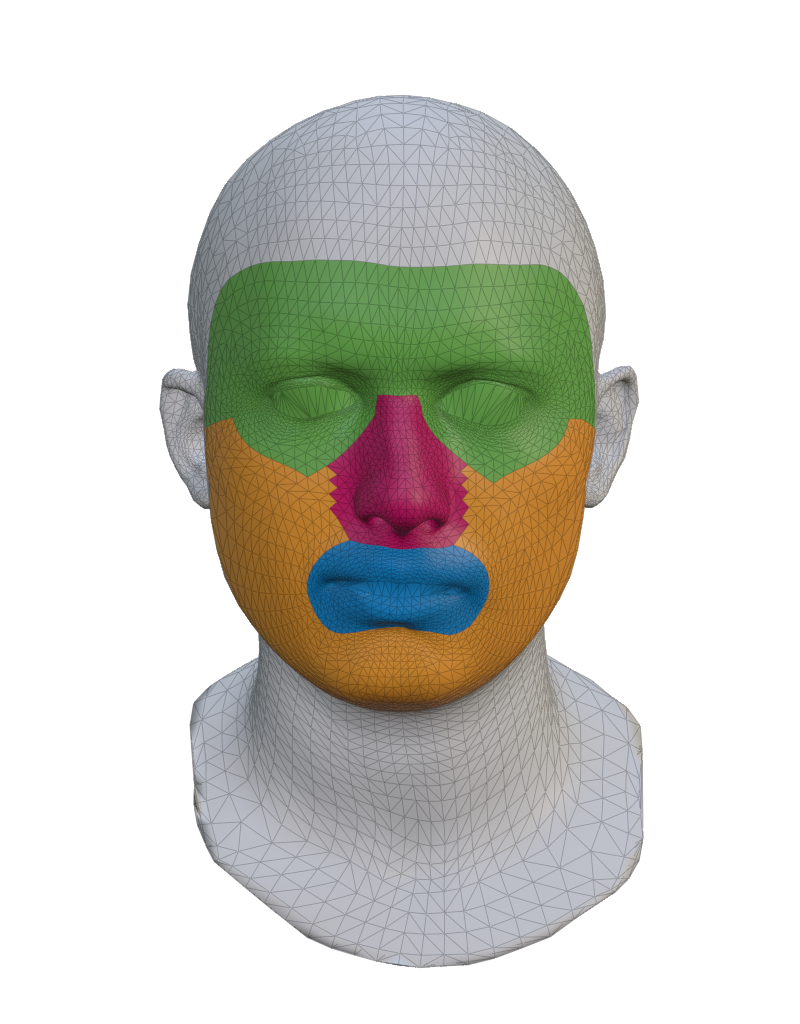}
    \caption{Visual representation of the forehead, nose, mouth, and cheek region masks used for our part-based evaluation. Masks do not overlap from one to another.}
    \label{fig:mask_regions}
\end{figure}

The evaluation is inspired by the REALY benchmark \cite{chai2022realy}. Four masks are considered in the Multiface topology, as illustrated in \cref{fig:mask_regions}. For a given mesh under evaluation, we first perform a rigid alignment of the evaluated regions from the ground truth to the generated mesh. For cheek and mouth, the rigid alignment is done with the combination of the mouth and cheek mask. After rigid alignment, we compute the mean vertex distance between the ground truth and the alignment mesh part.

\section{More qualitative results}
\begin{figure}
    \centering
    \includegraphics[width=1\linewidth]{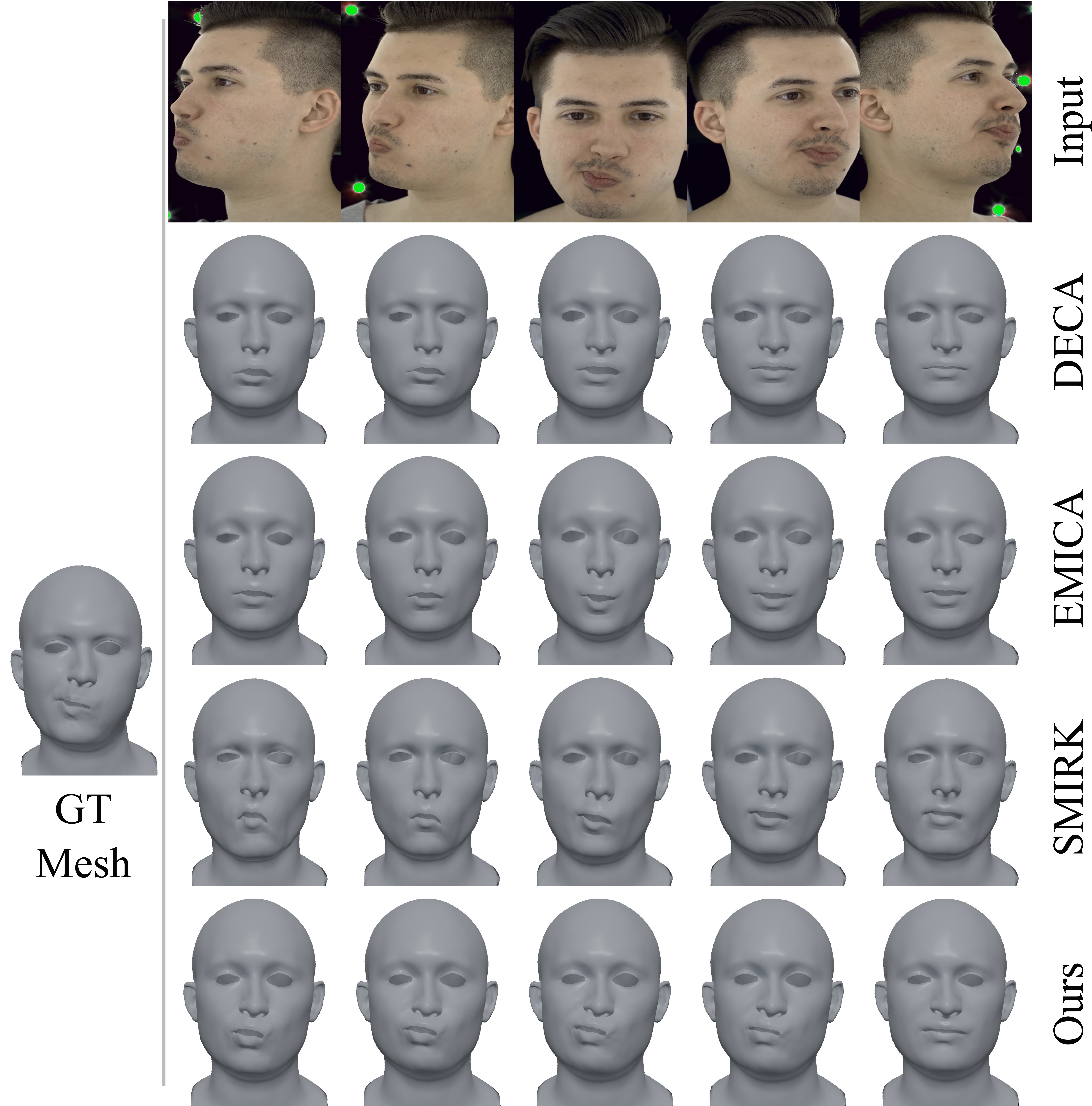}
    \caption{Additional captures on the proposed MultiREX benchmark}
    \label{fig:MultiREX_capture_0}
\end{figure}

\begin{figure}
    \centering
    \includegraphics[width=1\linewidth]{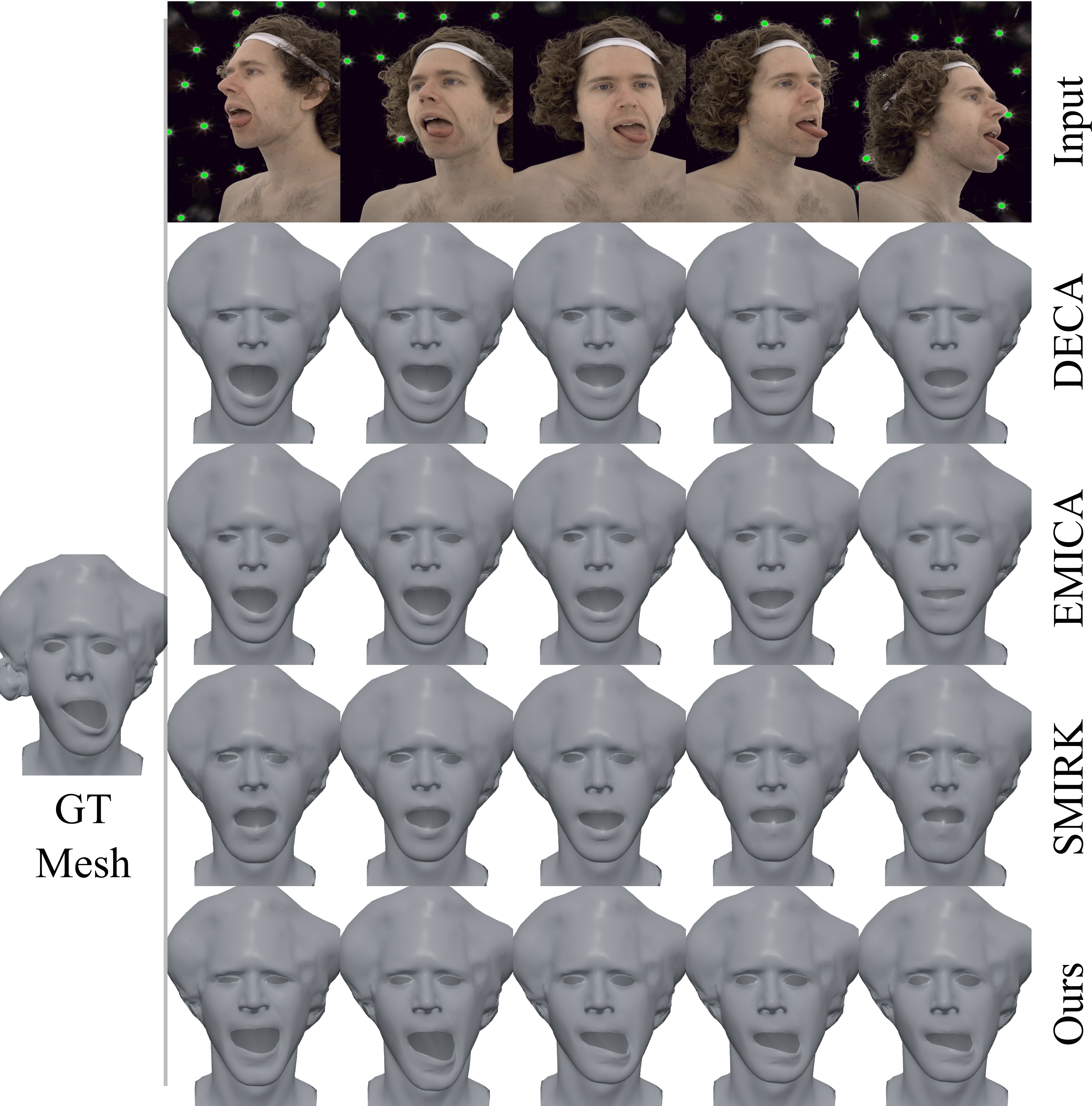}
    \caption{Additional captures on the proposed MultiREX benchmark}
    \label{fig:MultiREX_capture_1}
\end{figure}

\begin{figure}
    \centering
    \includegraphics[width=1\linewidth]{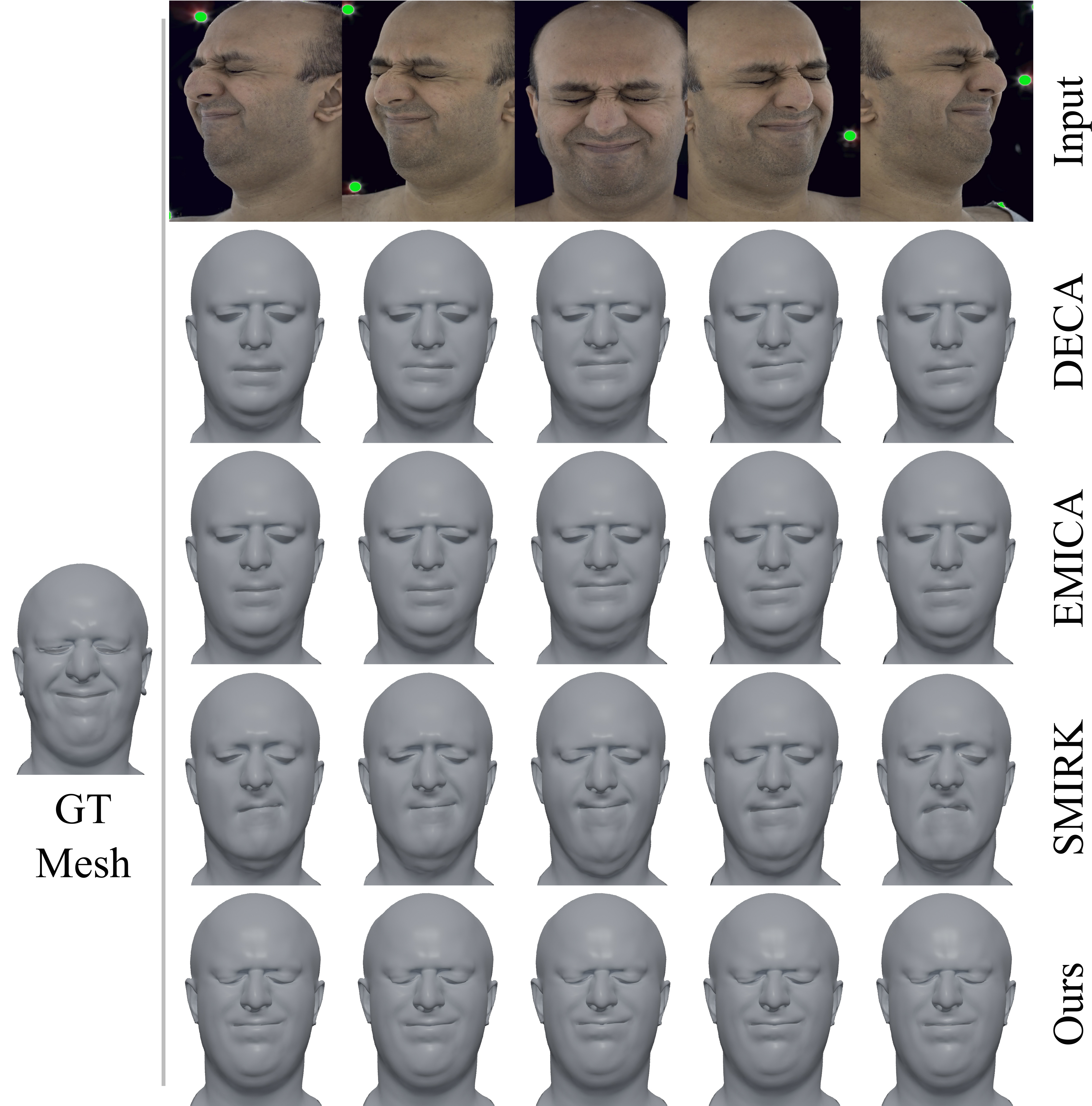}
    \caption{Additional captures on the proposed MultiREX benchmark}
    \label{fig:MultiREX_capture_3}
\end{figure}

\begin{figure}
    \centering
    \includegraphics[width=1\linewidth]{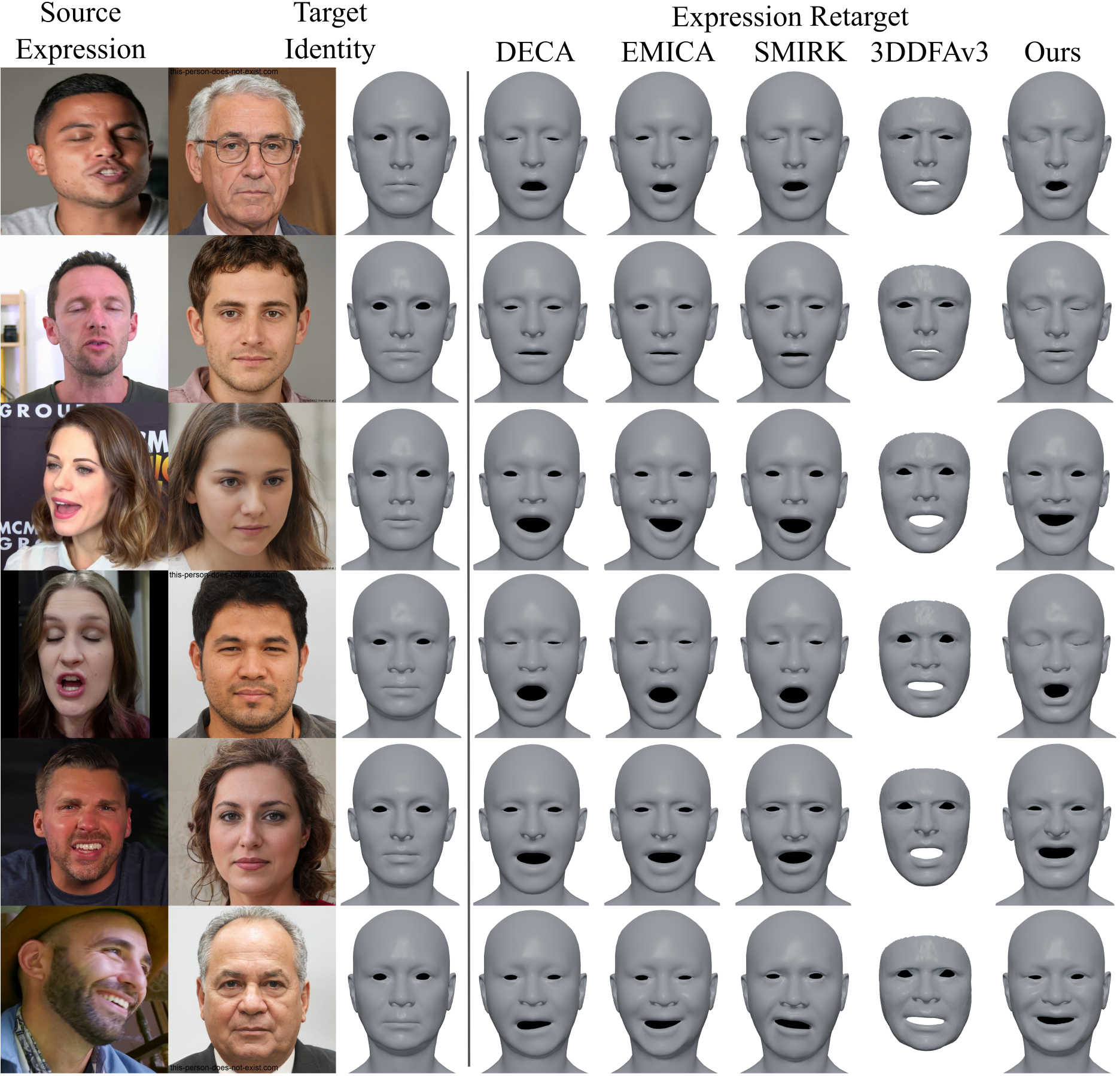}
    \caption{Additional retargeting on in-the-wild images}
    \label{fig:ITW_retargeting}
\end{figure}

\subsection{Comparison on MultiREX benchmark}
In this section, we show more visual comparison against state-of-the-art methods on the MultiREX benchmark.  Results are reported in \cref{fig:MultiREX_capture_0}, \cref{fig:MultiREX_capture_1} and \cref{fig:MultiREX_capture_3} for different subjects under different viewing angles. Our method shows robustness against side view changes and preserves better the subject's expression compared to other methods that generate less consistent expression over different views. 

\subsection{Comparison on in-the-wild images}
\cref{fig:ITW_retargeting}, we show more qualitative results for in-the-wild retargeting to other subjects and considering different source expressions. Our model is more faithful to the source expression in most scenarios while accounting for the face morphology and through its semantic expression model.  

\subsection{Challenging in-the-wild capture}
\begin{figure}
        \vspace{-1em}
        \centering
        \includegraphics[width=1\linewidth]{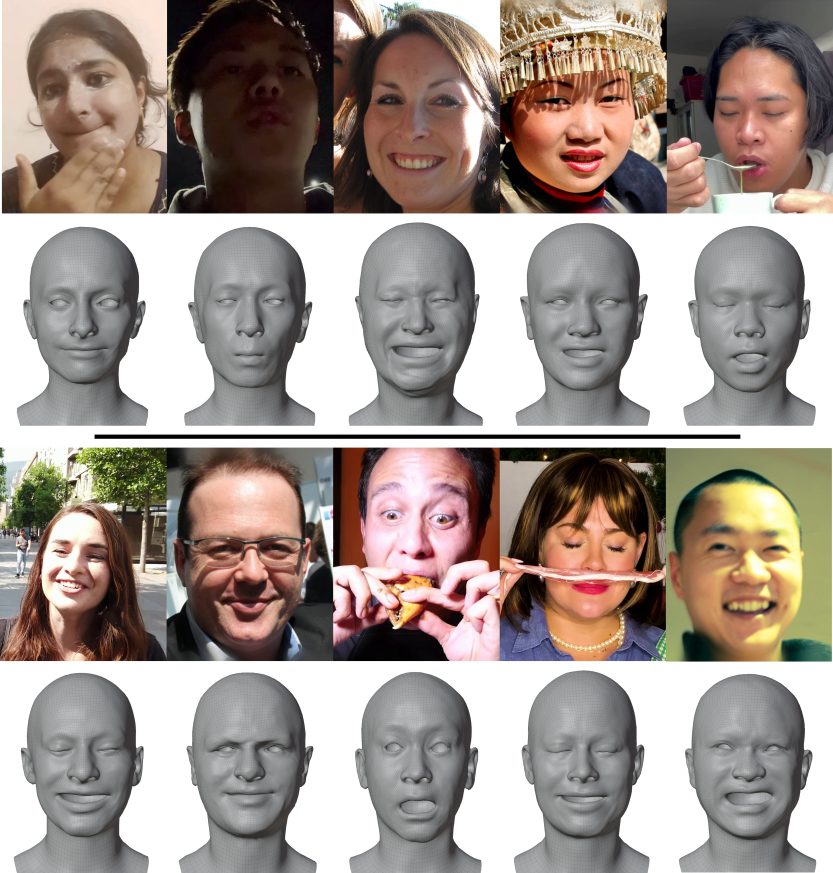}
        \caption{Results on challenging conditions}
        \label{fig:supp_examples}
        \vspace{-1em}
    \end{figure}

We complete the evaluation of our expression capture model with challenging in-the-wild captures and some failure cases \cref{fig:supp_examples}. Our model shows some  robustness to lighting conditions, 
but can fail in cases of external occlusions, and motion blur. Motion blur is particularly problematic in video frames, resulting in jittery capture. The overall robustness of our method is in part dependent on the quality of the real landmarks used for training, obtained using an off-the-shelf detector.

\section{Real-time processing} On a 2080Ti GPU (batch size of 1), the expression encoder takes 20.6 ms on average, and the mesh decoder takes 5.5 ms, for a total of 26.1 ms (38.3 FPS). This does not include bounding box detection.

\section{Synthetic dataset}
\begin{figure}
    \centering
    \includegraphics[width=1\linewidth]{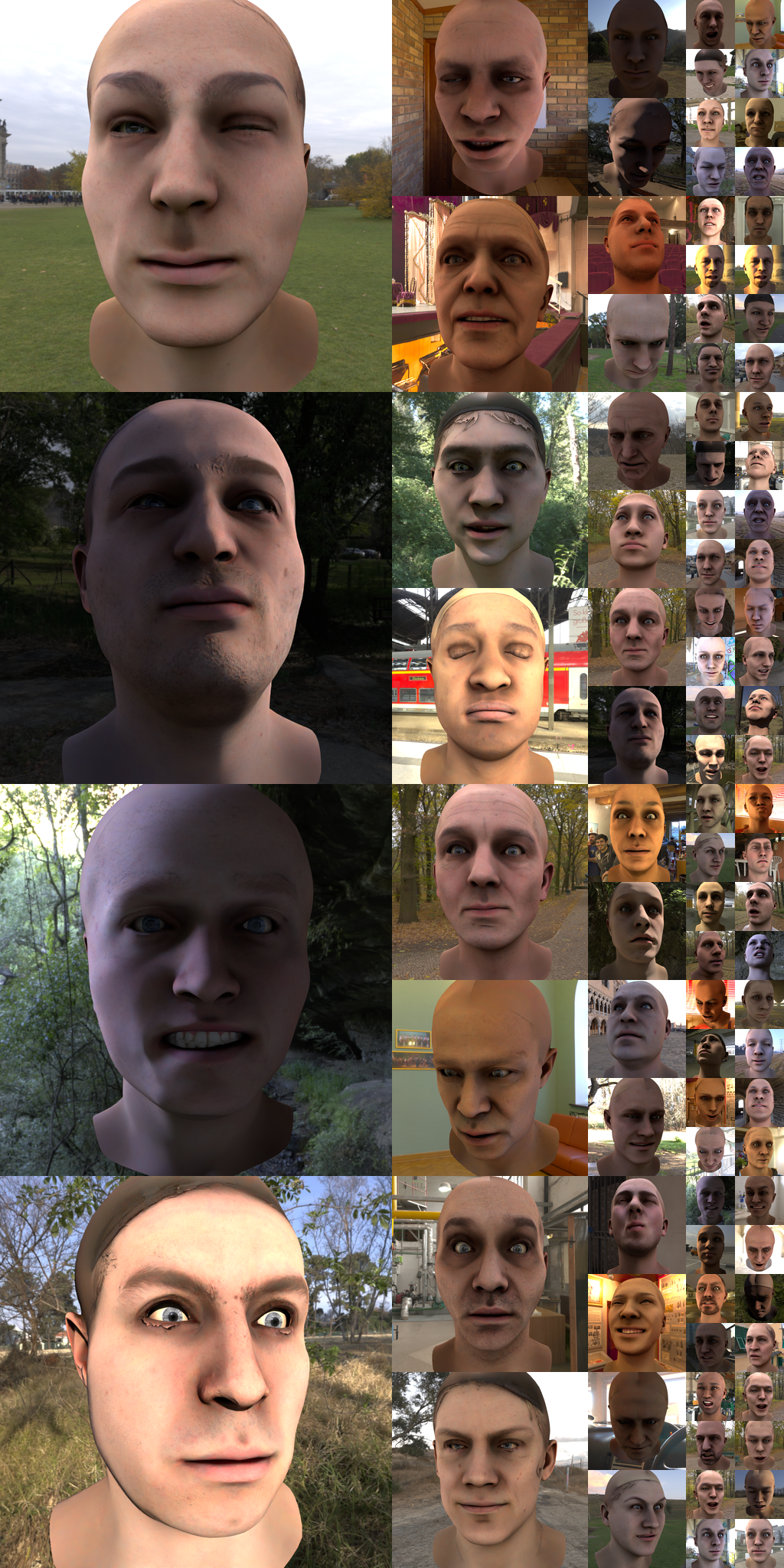}
    \caption{Random synthetic data samples.}
    \label{fig:synth_mosaic}
\end{figure}

\cref{fig:synth_mosaic} shows samples from our synthetic dataset used to train our expression capture model. 
Rendering is done using Blender, with the Cycles renderer. We use randomly selected environment maps\footnote{https://polyhaven.com}. We use the same meshes and textures for teeth and eyes for all subjects. They are placed procedurally based on the vertex positions of the eyelids and jaw.

We emphasize that the generation of our synthetic dataset does not require 3D modeling for hair, facial accessories, or clothes, contrary to \cite{wood2021fake}.

\end{document}